\pdfoutput=1

\documentclass[11pt]{article}
\usepackage{amsmath}
\usepackage[final]{acl}

\usepackage{hyperref}
\usepackage{multirow}
\usepackage{times}
\usepackage{latexsym}

\usepackage[T1]{fontenc}
\usepackage{booktabs}

\usepackage[utf8]{inputenc}

\usepackage{microtype}

\usepackage{inconsolata}

\usepackage{graphicx}
\usepackage{amsfonts}
\title{{\sc \textbf{PledgeTracker}}: A System for Monitoring the Fulfilment of Pledges}

\author{Yulong Chen, Zhenyun Deng, Andreas Vlachos \\
  University of Cambridge \\
  \texttt{\{yc632, zd302, av308\}@cam.ac.uk} \\\And
  Michael Schlichtkrull \\
  Queen Mary University London\\
  \texttt{m.schlichtkrull@qmul.ac.uk}
  \AND
  David Corney, Nasim Asl, Joshua Salisbury, Andrew Dudfield\\
  Full Fact\\
\texttt{\{firstname.lastname\}@fullfact.org}
  \\}

\begin{document}
\maketitle
\begin{abstract}
Existing methods simplify the pledge monitoring task into a document classification task, overlooking its dynamic temporal and multi-document nature. 
To address this issue, we introduce \textsc{PledgeTracker}, a system that formulates pledge monitoring as structured event timeline construction. 
\textsc{PledgeTracker} consists of three core components: (1) a multi-step evidence retrieval module; (2) a timeline construction module and; (3) a fulfilment filtering module, enabling us to capture the evolving nature of the task. 
We evaluate \textsc{PledgeTracker} in collaboration with professional fact-checkers in real-world workflows, showing its superior effectiveness over Google search
and \texttt{GPT-4o} with \texttt{web\_search}.


\end{abstract}

\section{Introduction}


Political pledges are commitments and governance plans made by political parties or candidates, especially during their election campaigns, which aim to promote their policies~\cite{costello2008election, dupont2019kind}.
Monitoring the fulfilment of pledges helps measure government performance, reinforcing transparency in democracy and accountability. 
However, this task typically requires fact-checkers to retrieve and analyse relevant documents regularly (e.g., daily or weekly)~\cite{duval2020citizens,fornaciari-etal-2021-will, sahnan2025llmsautomatefactcheckingarticle}, which is resource-intensive, motivating the need for automated systems.

\begin{figure}
    \centering
    \includegraphics[width=1\linewidth]{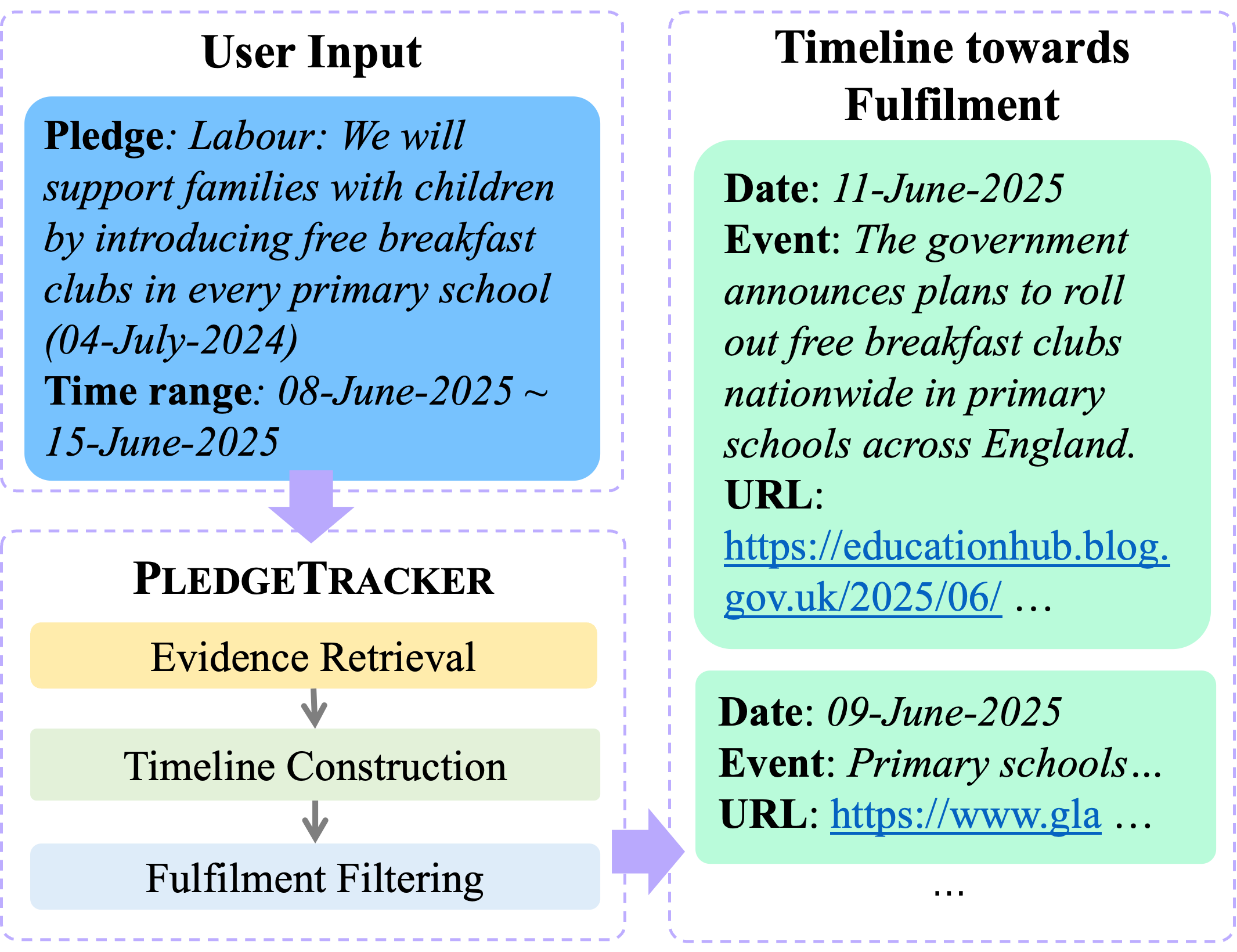}
    \caption{Overview of \textsc{PledgeTracker}.}
    \label{fig:overview}
\end{figure}


Recent work treats pledge monitoring as a \emph{document-level} classification problem~\cite{seki2024ml}, by identifying whether a single article supports a pledge or not, overlooking the dynamic and long-term nature of pledge fulfilment. 
A political pledge is a strategic commitment, which is usually fulfilled via a sequence of actions and milestones (e.g., ``\emph{build 100 new schools in the UK by 2027}'' materialises via local actions such as ``\emph{50 schools in England}'' or incremental milestones like ``\emph{30 schools by 2025}''). 
Furthermore, the pledge status is temporal and dynamic in nature.
It can evolve when new evidence emerges (e.g., the exit and re-entry into international agreements).
Thus the task requires collecting and reasoning over temporally distributed evidence from multiple documents.

These requirements distinguish pledge monitoring from conventional fact-checking~\cite{guo2022survey,schlichtkrull2023averitec,iqbal-etal-2024-openfactcheck}.
Although fact-checking also collects evidence from multi-document, it typically focuses on verifying whether a claim is supported by evidence \textit{before} when the claim was made \cite{konstantinovskiy2021claim}.
Thus, the verdict is unlikely to change as those claims are about facts or knowledge that have already happened, except for corrections due to errors. 
In contrast, pledge monitoring aims to track how the fulfilment of a pledge evolves. 
Moreover, unlike the static labels of fact-checking output, pledge monitoring requires the output to reflect incremental progress over time. 
As such, the needs of end-users go beyond static labels, calling for structured, time-aware output.

To address these issues, we introduce \textsc{PledgeTracker}, a retrieval augmented generation (RAG)-based system for monitoring the fulfilment of political pledges by extracting timelines from online documents.
As shown in \autoref{fig:overview}, \textsc{PledgeTracker} consists of three core components in a multi-step framework:
(1) an {evidence retrieval} module collects and identifies relevant documents through multi-step retrieval; 
(2) a {timeline construction} module identifies and extracts key event descriptions and their timestamps from multiple relevant documents;
(3) a {fulfilment filtering} module determines relevant events, and assembles them into a temporally-structured timeline~\cite{hu2024moments}. 
For the development of the latter, we construct an annotated dataset covering 1,559 event descriptions across 50 pledges, where each event is labelled regarding its relevance to fulfilment.
To demonstrate the effectiveness, we evaluate \textsc{PledgeTracker} in collaboration with professional fact-checkers from Full Fact, in their real-life workflows where evidence continuously evolves. 
{Our system achieves 0.64 $F_1$ in identifying fulfilment events in a real-world evaluation.}
Moreover, our further analysis finds \textsc{PledgeTracker} to be more accurate in retrieving useful evidence URLs (0.78 $F_1$) than Google Search (0.23 $F_1$) and \texttt{GPT-4o} with \texttt{web\_search} (0.03 $F_1$), both of which are part of the modules in \textsc{PledgeTracker}. Qualitative feedback suggests that \textsc{PledgeTracker} brings useful events to the attention of the fact-checkers that would have otherwise been missed.
%
We publicly release \textsc{PledgeTracker}\footnote{\url{https://huggingface.co/spaces/PledgeTracker/Pledge_Tracker}} and our annotation to facilitate the task of pledge monitoring.

\section{Pledge Monitoring}
Drawing inspiration from fact–checking organisations like Full Fact’s Government Tracker,\footnote{\url{https://fullfact.org/government-tracker/}} pledge monitoring refers to the task of fulfilling promises with actions, i.e., when, how, and to what extent those promises are being fulfilled.
We formulate this task as constructing an event timeline that reflects the progress regarding a pledge. 

Formally, given a pledge $p = (p_s, p_d, p_g, p_c)$, where $p_s$ is the pledge speaker (e.g., a political party such as \textit{Labour}), $p_d$ is the pledge date (i.e., when it is made), $p_g$ is the geographic scope (e.g., \textit{the UK}), and $p_c$ is the pledge claim (e.g., ``\textit{We will ban trail hunting}''), and a monitoring time range $r = (r_s, r_e)$, where $r_s$ and $r_e$ are the start date and end date, respectively, the system $\mathcal{S}$ is asked to generate a timeline $T$:
\begin{equation}
  T=\mathcal{S}(p, r),
\end{equation}
where $T$ is the timeline (possibly empty if no progress has been made).
For a non-empty $T=\{(e, t, url)\}$, each event description $e_i$ is associated with a timestamp $t_i$ and its source URL $url_i$, with the full set sorted in order, i.e., for all $i < j$, we have either $t_i \leq t_j$ (chronological) or $t_i \geq t_j$ (reverse chronological).
%
%
Timeline $T$ captures incremental progress and setbacks over time.

\begin{figure*}
  \centering
  \includegraphics[width=\linewidth]{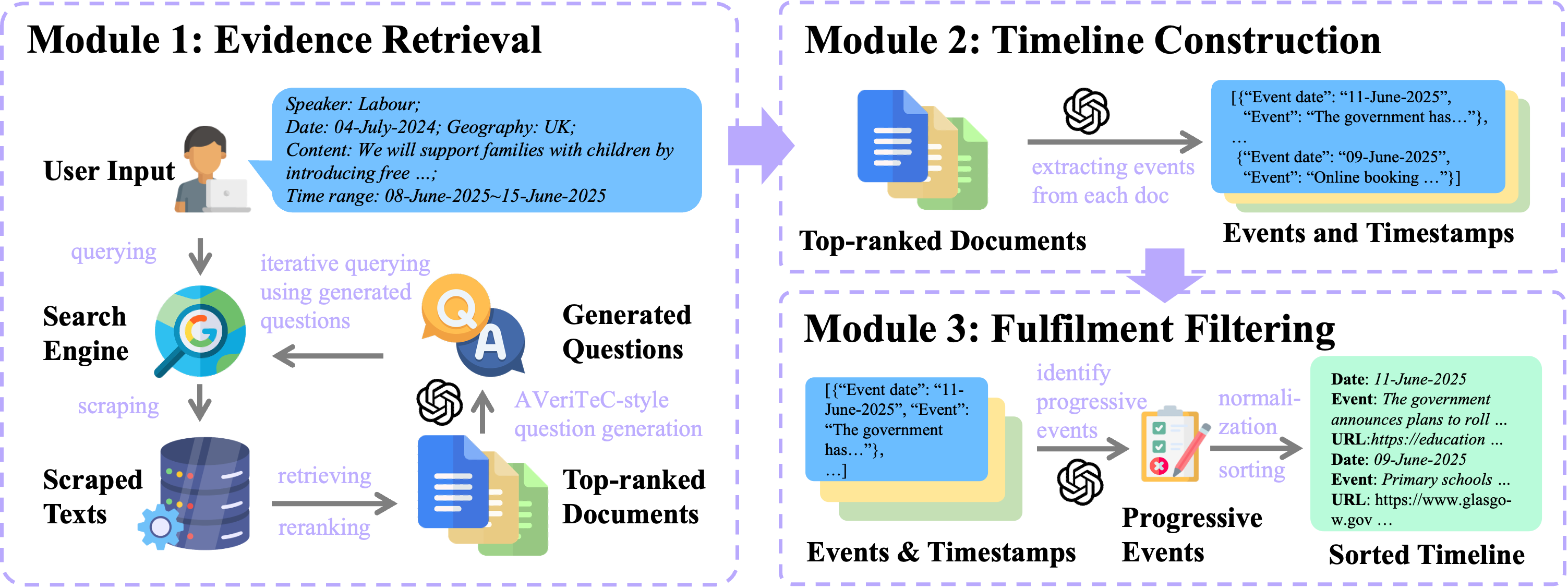}
  \caption{The architecture and workflow of \textsc{PledgeTracker}.}
  \label{fig:system}
\end{figure*}

\section{{\sc \textbf{PledgeTracker}}}\label{sec:pledge_tracker}

As shown in \autoref{fig:system}, \textsc{PledgeTracker} is a RAG-based system
consisting of three modules: an evidence retrieval module $\mathcal{R}$, a timeline construction module $\mathcal{T}$, and a fulfilment filtering module $\mathcal{F}$, i.e., $\mathcal{S}=\{\mathcal{R, \mathcal{T}, \mathcal{F}}\}$.
Given a pledge and the time range, we first collect a set of documents using the evidence retrieval module: $D=\mathcal{R}(p, r)$.
Then, based on the retrieved documents and the pledge, the timeline construction module extracts all possible events and their timestamp: $E=\mathcal{T}(D, p)$.
Finally, the fulfilment filtering module selects the subset of events most useful to monitor the pledge, producing the final timeline: $T = \mathcal{F}(E, p, r)$.
The subsequent subsections provide a detailed description of the corresponding modules. 

\subsection{Evidence retrieval}\label{sec:deep_retrieval}

Following recent work on evidence retrieval that uses a multi-round retrieval strategy~\cite{liao2023muser,yang2024rag, liu2024information}, \textsc{PledgeTracker}'s retrieval component is progressively expands and refines the document set $D$ in multiple rounds of interaction and question-guided augmentation.

Given a pledge $p$ and a target monitoring time range $r$, we first perform an {initial web search} using Google custom search API.\footnote{\url{https://developers.google.com/custom-search/}} In particular, we construct a query string such as \textit{``Labour: We will ban trail hunting (04-Jul-2024)''}, conditioned by the geographic scope $p_g$ and the date range $(r_s, r_e)$. 
As these results can often be sparse or incomplete, we further extract key noun phrases (e.g., ``\textit{trail hunting}'') from the pledge content $p_c$ using \texttt{spaCy}\footnote{\url{https://spacy.io/}} as additional search queries.
Given the retrieved URL results, we obtain the corresponding textual documents using \texttt{trafilatura}~\cite{barbaresi-2021-trafilatura}, a library for web crawling and text extraction.

To guide deeper retrieval, we further incorporate question-driven augmentation based on retrieved evidence.
Following \citet{yoon-etal-2024-hero}, we first generate a set of hypothetical documents, which simulate possible evidence. 
We then use those hypothetical documents to retrieve sentence-level evidence from the scraped texts using \texttt{bm25}, and re-rank the evidence based on their semantic similarity computed with \texttt{SFR-Embedding-2\_R}.\footnote{\url{https://huggingface.co/Salesforce/}}
For each top-ranked evidence, we generate the corresponding clarification question that explicitly targets different aspects of the pledge (e.g., \textit{``Is Labour planning to implement a central reporting mechanism for reporting potential animal welfare offences?''}). These questions are then used as new search queries for the next round of retrieval.
Both the hypothetical document generation and question generation are performed by \texttt{Llama-3.1-8B-instruct}~\cite{grattafiori2024Llama} using in-context learning (ICL) examples from AVeriTeC~\cite{schlichtkrull2023averitec}.
The details can be found in Appendix~\ref{append:prompt_question_generation}.

Finally, after multiple rounds of retrieval,
the evidence retrieval module returns a set of top-ranked evidence. 
We then collect and deduplicate the document texts and corresponding URLs to construct the final $D$ for the timeline construction module as described in the next subsection.




\begin{figure*}[ht]
    \centering
    \includegraphics[width=0.9\linewidth]{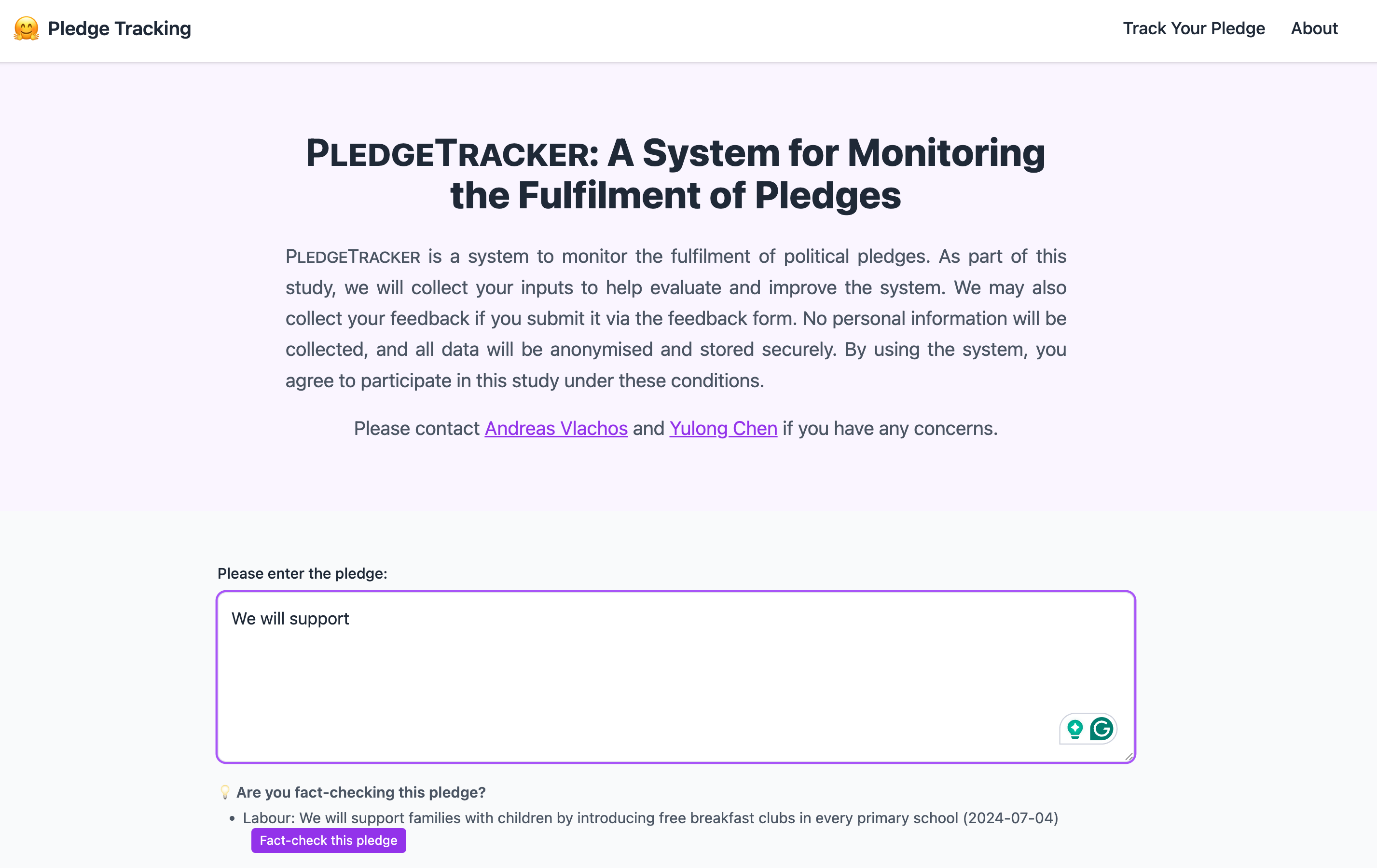}
    \caption{The user input interface of \textsc{PledgeTracker}.}
    \label{fig:input_interface}
\end{figure*}


\subsection{Timeline Construction}\label{sec:event_timestamp}

Rather than relying on predefined schemas~\cite{minard2015semeval}, we adopt a generative extraction approach using \texttt{GPT-4o}~\cite{hurst2024gpt}, which allows for more flexible identification of events~\cite{gao2023exploring,chen2024large,qorib2025just}.

In particular, we prompt the model using few-shot ICL examples consisting of document–event pairs that we annotated manually, and constrain the model output to follow the JSON format.
Given a pledge $p$ and each document $d_i \in D$, we construct a prompt that includes the document’s metadata (e.g., title and publication date), the article body, and the pledge text, in order to generate relevant event descriptions (e.g., \textit{``A petition is rejected because there is already a similar petition about banning trail hunting.''}).
Moreover, since event timestamps mentioned in the text may be expressed in various terms (e.g., publication date: \texttt{08-Jul-2024}, event temporal-related phrase: \texttt{``two days ago''}), we prompt the model through ICL to generate the corresponding absolute date (e.g., \texttt{06-Jul-2024}) if possible, or a relative date (e.g., \texttt{Last month (relative to 01-Jul-2024})).
The details can be found in Appendix~\ref{append:prompt_eevent_generation}.

After processing all documents from $D$, we further sort the events by their dates.
We normalise the timestamps using a rule-based parser that handles a wide range of temporal expressions (e.g., locating ``\texttt{Autumn 2023}'' into ``\texttt{01-09-2023}'').
Finally, this module returns a set of candidate events $E$.




\subsection{Fulfilment Filtering}\label{sec:process_identification}
In practice, we find that not all events in $E$ are informative or relevant to monitoring the fulfilment of the pledge.
Although they are extracted from top-ranked documents, many events provide only contextual or background information (e.g., \textit{What does the pledge mean?}), rather than concrete progress to fulfilling the pledge (\textit{What progress has been made?}).\footnote{\url{https://fullfact.org/government-tracker/hillsborough-law-candour-duty/}}
For example, while the event ``\textit{Critics claim trail hunting is being used as a `smokescreen' for illegal fox hunting activities}'' is related to ``\textit{trail hunting}'', it does not contain any useful information about the actions that were taken.
To address this,
we developed the fulfilment filtering module $\mathcal{F}$ to filter the events to be included in the timeline.


\begin{figure*}[ht]
    \centering
    \includegraphics[width=0.9\linewidth]{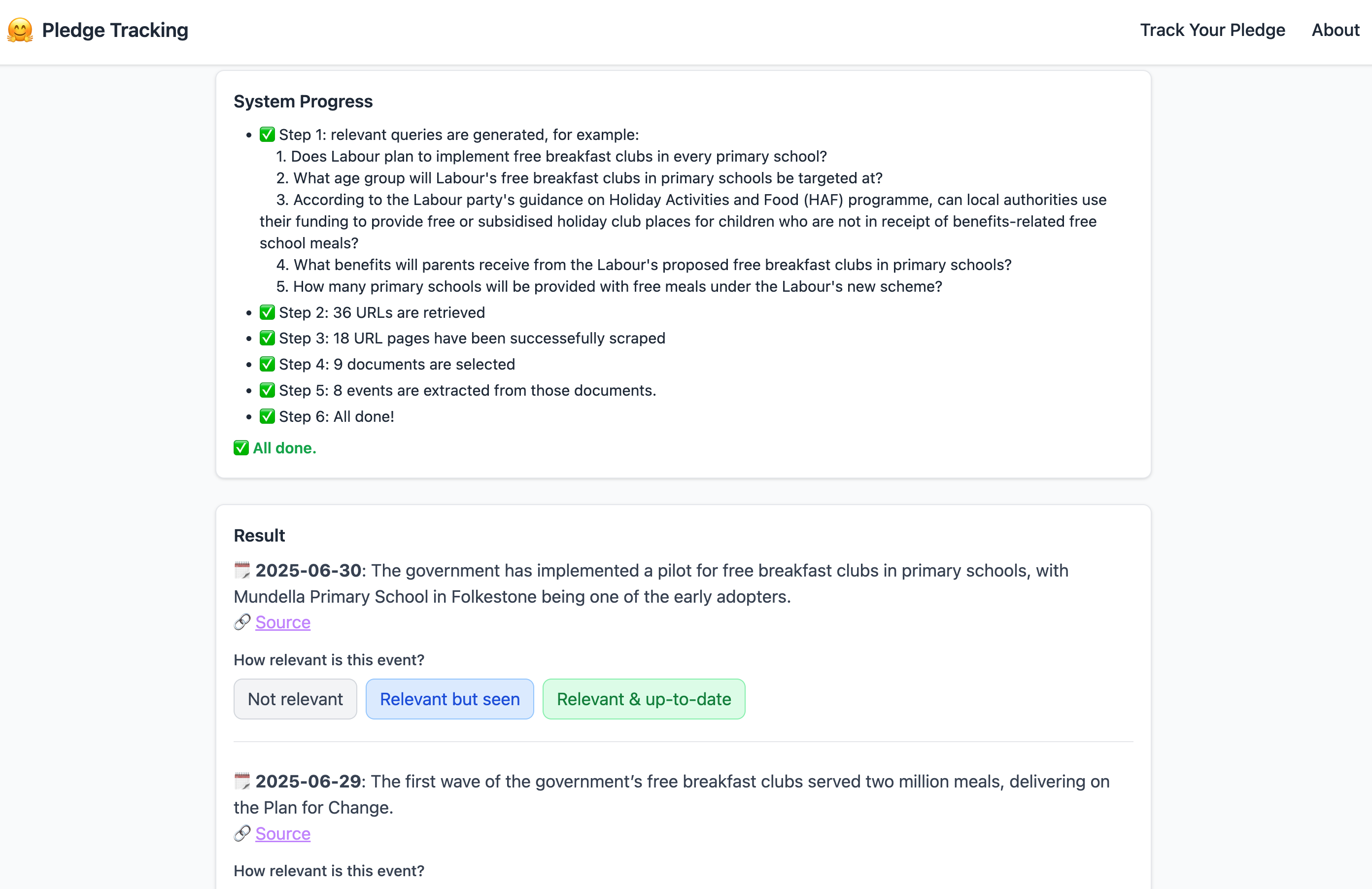}
    \caption{The output timeline interface of \textsc{PledgeTracker}.}
    \label{fig:output_interface}
\end{figure*}

To support the fulfilment filtering, we construct a dataset focusing on the task.
We begin with a set of 50 pledges selected from FullFact government tracker, which are from the Labour Party's manifesto for the 2024 UK general election.
We then use the \textsc{PledgeTracker} (without fulfilment filtering) to retrieve all potentially related events from the time each pledge was made (starting on 4 July 2024) up to the time when the timeline was generated (March 2025).
For each pledge, a professional fact-checker, who was familiar with it, examined the generated timeline and evaluated whether each event and its timestamp were useful or not, with the help of the corresponding URL.
In particular, we define an event and its timestamp as \textit{useful} in assessing fulfilment if it (1) is factually consistent with the source document, (2) contains a correctly inferred timestamp, and (3) contributes to the fulfilment of the pledge.
If any of these criteria are not met, the event is labelled as \textit{not useful}.
In total, we collect 1,559 annotated instances, where each instance consists of a pledge, an event description, a timestamp, the original URL, and a binary usefulness label.
In particular, our analysis shows that only 26.63\% of them are useful in monitoring the fulfilment of the corresponding claims, which demonstrates the necessity of fulfilment filtering.

During testing, given each $e_i$ from $E$, we ask \texttt{GPT-4o} to label each extracted event as either \textit{useful} or \textit{not useful} in assessing fulfilment using ICL examples from our annotation.
The resulting timeline provides a clear and interpretable progression of pledge fulfilment over time.
The details can be found in Appendix~\ref{append:prompt_relevant_event}.

\section{User Interface Design}\label{sec:interface}


We build the \textsc{PledgeTracker} demo system on Hugging Face Space 
(Nvidia A100) using Flask.
%
Using the interface, users enter a pledge, specify the speaker, pledge date, and time range, and initiate the system by clicking the ``\textit{Let's track!}'' button (\autoref{fig:input_interface}).

Once the input data is submitted, \textsc{PledgeTracker} starts the multi-stage pipeline as detailed in \S\ref{sec:pledge_tracker}.
The system will start collecting evidence, generating the timeline, and identifying fulfilment events using ICL instances from our annotation,
showing relevant status updates (\autoref{fig:output_interface}).
Finally, \textsc{PledgeTracker} presents the timeline, where each event is associated with an event date, an event description, and the original source link.
To support iterative refinement for analysis and future work, the demo system enables users to provide feedback on the usefulness of each event.

\textsc{PledgeTracker} also supports matching pledges against previously checked ones.
When the user enters a new pledge, the system automatically searches for similar pledges among the pledges already checked by the system, using \texttt{TF-IDF} and shows the top suggestions based on their similarities.
For suggested pledges, the system will re-use previously retrieved results (from an initial web search) to accelerate the process and enable more accurate fulfilment filtering by selecting corresponding annotated data.






\section{Experiments}

We perform two kinds of quantitative evaluation: offline, using our annotated data, and in real-world use with professional fact-checkers. 
In particular, we first demonstrate offline the effectiveness of fulfilment filtering
(\S\ref{sec:exp_fulfilment}), and then evaluate the full \textsc{PledgeTracker} in real-world use (\S\ref{sec:real_time_exp}) and show its comparison with existing tools (\S\ref{sec:comparioson_existing}).
We further present qualitative analysis in \S\ref{sec:qualitative}.

\subsection{Effectiveness of Fulfilment Filtering}\label{sec:exp_fulfilment}

As described in \S\ref{sec:process_identification}, we collect 1,559 instances for fulfilment filtering, which are divided into training (949), development (249) and test (361) sets based on pledges.
\autoref{tab:data_stat} shows their statistics.
We note the distribution difference across data splits due to fulfilment varying across pledges. 
\begin{table}
\centering
    \small
    \begin{tabular}{l|ccc}
    \toprule
         & \textbf{Train} & \textbf{Dev} & \textbf{Test}   \\
         \midrule
         useful (\%) & 20.86 & 33.33 & 37.12\\
         non-useful (\%) & 79.14 & 66.67 & 62.88 \\
         event/pledge & 43.14 & 24.90 & 20.06 \\
         \bottomrule
    \end{tabular}
    \caption{Statistics for the fulfilment filtering annotation.}
    \label{tab:data_stat}
\end{table}%
We conduct experiments using three models: (1) \textsc{RoBERTa}-large~\cite{liu2019roberta} with full-parameter fine-tuning; (2) \texttt{Llama-3-8B} \cite{grattafiori2024Llama} trained using instruction-based LoRA tuning \cite{hu2022lora} and; (3) \texttt{GPT-4o} with ICL prompting. 
Given a pledge and an associated event, each model is asked to assign a binary label indicating whether the event is useful in assessing fulfilment. 

\begin{table}[]
    \centering
    \small
    \begin{tabular}{l|ccc}
    \toprule
    & \textbf{P} &\textbf{R} &\textbf{\boldmath $F_1$}\\
    \midrule
        \textsc{RoBERTa} & 0.517 & 0.224 & 0.313   \\
        \texttt{Llama} & 0.544 & 0.507 & 0.525 \\
        \texttt{GPT-4o} & 0.509 & 0.836 & 0.633\\
        \bottomrule
    \end{tabular}
    \caption{Results on fulfilment filtering.}
    \label{tab:development}
\end{table}

\begin{table*}[h]
  \centering
  \small
  \begin{tabular}{l|ccc|ccc|c}
    \toprule
    \multirow{2}{*}{\textbf{System}} & \multicolumn{3}{c|}{\textbf{Pledge-level}} & \multicolumn{3}{c|}{\textbf{URL-level}} & \multirow{2}{*}{\textbf{Novelty}} \\
     & P & R & $F_1$ & P & R & $F_1$ & \\
    \midrule
    \textsc{PledgeTracker}       & {0.83} & {0.74} & {0.76} & {0.93} & {0.68} & {0.78} & {36} \\
    Google Search              & 0.32 & 0.08 & 0.12 & 0.50 & 0.15 & 0.23 & 5 \\
    \texttt{GPT-4o} with \texttt{web\_search}                 & 0.08 & 0.01 & 0.01 & 1.00 & 0.02 & 0.03 & 1 \\
    \bottomrule
  \end{tabular}
  \caption{Overall retrieval performance. Pledge-level: results first averaged per pledge, then averaged. 
  URL-level: results averaged across all URLs. Novelty: the number of unique useful URLs retrieved by a system.}
  \label{tab:comparison_existing_tool}
\end{table*}

As shown in \autoref{tab:development}, \texttt{GPT-4o} achieves the best performance, with an $F_1$ score of 0.633. 
It suggests that, compared with \textsc{RoBERTa} and \texttt{Llama}, \texttt{GPT-4o} is better at capturing potential fulfilment signal.
The main challenge lies in the imbalanced data distribution of the pledge monitoring data. 
As mentioned, the fulfilment events can be sparse in the real world, while most events lack concrete evidence of progress (c.f. \autoref{tab:data_stat}). 

\subsection{Evaluation in Real-world Use}\label{sec:real_time_exp}
After deploying the full version of \textsc{PledgeTracker}, we evaluate the system in a \textit{real-world} setting with Full Fact fact-checkers.
{In particular, our evaluation was conducted from 12 June to 08 September 2025, monitoring 68 pledges from the Labour Party's 2024 UK election manifesto.}
Each timeline is generated over a time range of the past 7 days. 
{As some pledges were monitored multiple times at different times in the evaluation period, we collected 113 timelines in total.}
Two professional fact-checkers (paper co-authors Nasim Asl and Joshua Salisbury), who were responsible for the corresponding pledges in their daily work, evaluate the usefulness of each event, using the criteria described in \S\ref{sec:process_identification}.
We continue to present \textit{all} candidate events, including both those retained and those filtered out, to the fact-checkers.
This setup enables a direct comparison between the \textsc{PledgeTracker}'s filtering decisions and human judgments.
During the evaluation, the fact-checkers select one of three labels: \texttt{not\_relevant}, \texttt{relevant\_seen}, and \texttt{relevant\_update}. 
The label \texttt{relevant\_update} indicates that the event is new to the fact-checkers and useful for fulfilment tracking, \texttt{relevant\_seen} means that the event is useful and temporally appropriate, and meanwhile, fact-checkers already know about it. 
We therefore treat both \texttt{relevant\_seen} and \texttt{relevant\_update} as \textit{useful} in our evaluation, since our goal is to assess whether the system can accurately surface relevant fulfilment evidence, regardless of whether the annotator had seen it from other sources.
{In total, 513 events were evaluated across 68 timelines.}

\begin{table*}[h]
\centering
\small
\begin{tabular}{p{0.1cm} p{5.6cm} p{1cm} p{7.5cm}}
\toprule
\textbf{ID} & 
\multicolumn{1}{c}{\textbf{Pledge claim}} & 
\multicolumn{1}{c}{\textbf{Date}} & 
\multicolumn{1}{c}{\textbf{Event description, timestamp and URL}} \\
\midrule
1 & Labour will end the VAT exemption and business rates relief for private schools & 2025-06-13 & Private school families lost their High Court challenge against the Government over the VAT policy on fees. 2025-06-13. [\href{https://www.telegraph.co.uk/news/2025/06/13/private-school-families-lose-challenge-labour-vat-raid/}{URL}] \\
\midrule
2 & Labour will capitalise Great British Energy with £8.3 billion, over the next parliament & 2025-06-11 & The government is delivering a new generation of publicly owned clean power. Great British Energy and Great British Energy–Nuclear will together invest more than £8.3 billion over the SR in homegrown clean power. 2025-06-11. [\href{https://www.gov.uk/government/publications/spending-review-2025-document/spending-review-2025-html}{URL}] 
\\
\bottomrule
\end{tabular}
\caption{Events that led to updates in Full Fact's pledge pages. The Date here refers to when the monitoring was requested. The time range is set to the past 7 days. We attach the hyperlink (URL) for reference.}
\label{tab:pledge-case}
\end{table*}

{Generally, \textsc{PledgeTracker} achieves 0.764 precision, 0.553 recall and 0.641 $F_1$, demonstrating that it can identify fulfilment events with reasonably high performance in a real-world setting.}
Compared to the offline results in \S\ref{sec:exp_fulfilment}, the full system shows higher precision.
This can be partly because the full system benefits from using the full annotation set for ICL prompting.
Meanwhile, recall slightly decreases, which can be because the time range (past 7 days) is narrower, resulting in sparser fulfilment.
{In particular, for 513 events, we manually identify 152 fulfilment events (29.63\%), which is lower than in the offline evaluation (37.12\%).}

\subsection{Comparison with Existing Tools}\label{sec:comparioson_existing}

We compare \textsc{PledgeTracker} with two other tools that are often used for pledge monitoring: (1) Google Search and; (2) \texttt{GPT-4o} with \texttt{web\_search}.
In particular, we collect 13 pledge monitoring requests {(from 12 June to 22 June 2025)} from the evaluation in \S\ref{sec:real_time_exp} that received at least one fulfilment event according to the fact-checker's judgment. 
We use the aforementioned two tools to return top-ranked evidence (Appendix~\ref{append: gpt_4o_for_evidence_retrieval}), and ask fact-checkers to evaluate them.
Since they cannot directly return timelines, the evaluation focuses on \textit{whether the retrieved URLs include events useful in assessing fulfilment}.
For each pledge monitoring request, we first pool all URLs returned by the three systems, remove duplicates, and have professional fact‐checkers label each URL.
We take all \textit{useful} URLs for a given request as the ground truth set and evaluate their performance as shown in \autoref{tab:comparison_existing_tool}.

Overall, \textsc{PledgeTracker} retrieves 68\% of all manually identified evidence with 0.93  precision and 0.78 $F_1$ at the URL level.
It also contributes 36 unique, useful URLs that other systems fail to find. 
Compared to Google Search (0.15 recall), \textsc{PledgeTracker} benefits from the question-driven iterative retrieval using question generation, which aligns with findings from the AVeriTeC~\cite{schlichtkrull2023averitec,schlichtkrull2024automated}.
It is worth noting that \textsc{PledgeTracker} has higher precision than Google Search (0.50), indicating the effectiveness of our other modules.
Moreover, \texttt{GPT-4o} shows very poor performance in this task (0.03 $F_1$ at URL level).
In our evaluation, we find that \texttt{GPT-4o} is less sensitive to temporal constraints. 
In particular, although \texttt{GPT-4o} returns 61 URLs in total, only 1 is within the correct time range. 

\subsection{Qualitative Feedback}\label{sec:qualitative}
The two Full Fact fact-checkers who conducted the human evaluation in \S\ref{sec:real_time_exp} also provided some qualitative feedback.
From their feedback and specific examples as shown in \autoref{tab:pledge-case}, we observe certain scenarios where the system has been helpful.

First, \textsc{PledgeTracker} captures useful events that may otherwise be overlooked.  
In \autoref{tab:pledge-case} case 1, it alerted fact-checkers to news that had not gained much coverage in the media, a High Court ruling. 
Although this event did not change the verdict of the pledge, it led to an update to the pledge page on the removal of the VAT exemption for private schools, as the page previously said the appeal was taking place.
Second, \textsc{PledgeTracker} assists in timely event identification.
In \autoref{tab:pledge-case} case 2, \textsc{PledgeTracker} found that the investment would be split between Great British Energy and Great British Nuclear, on the same day the government's 2025 Spending Review was released.
This early signal enabled them to update the pledge page promptly and contact the UK government for further clarification. 
Third, \textsc{PledgeTracker} helps surface legislative and political signals that inform future developments.
Fact-checkers found \textsc{PledgeTracker} could highlight the names of bills and draft legislations associated with pledges, and trace their mentions across time in official communications. 
For example, it surfaces passing remarks by politicians, indicating when legislation or announcements could be expected, which was not previously captured through routine monitoring. 
Overall, they reported that \textsc{PledgeTracker} greatly contributes to their workflow.





In addition to these strengths, fact-checkers also noted occasional hallucinations in the event descriptions, for example, the generated events can be inconsistent with the source documents. 
To mitigate this known limitation of LLMs~\cite{zhang2023siren,chen2024see}, \textsc{PledgeTracker} is designed to explicitly include source URLs for each event, allowing fact-checkers to verify the underlying evidence when necessary. 




\section{Conclusion and Future Work}
We presented \textsc{PledgeTracker}, the first end-to-end system that formulates pledge monitoring as the construction of temporally ordered timelines. 
By iteratively collecting evidence from online, with generative timeline construction and fulfilment filtering, \textsc{PledgeTracker} captures incremental evidence and generates more interpretable outputs.
We integrated the system into professional fact-checkers' real-life workflows, and found
\textsc{PledgeTracker} achieved an $F_1$ of 0.641 in identifying fulfilment events. 
Our further comparison with Google Search and \texttt{GPT-4o} with \texttt{web\_search}, demonstrating the superior performance of \textsc{PledgeTracker} for pledge monitoring.


\section*{Limitations}

The limitations of \textsc{PledgeTracker} can be stated from four perspectives.
First, \textsc{PledgeTracker} is built on the basis that pledges have already been identified and normalised, and therefore it does not address the task of automatically extracting and decontextualising pledges from manifestos~\cite{deng-etal-2024-document, panchendrarajan2024claim}.
Second, our evaluation focuses on pledges from UK political parties.
However, its effectiveness in other linguistic or institutional contexts remains to be further explored~\cite{zhang2024we,turk2025clac}. 
Third, our evidence retrieval relies heavily on the Google Custom Search API, limiting its evidence coverage with potential ranking bias, and quota constraints.
Fourth, due to the limited resources, we could not perform large-scale training and thus use small models and LLM APIs for implementing \textsc{PledgeTracker}.

\section*{Ethical Considerations}

Our work involves human annotation and evaluation as stated in \S\ref{sec:process_identification}, \S\ref{sec:real_time_exp}, and \S\ref{sec:comparioson_existing}.
These two annotators are professional fact-checkers and the co-authors of this paper.
Their background information is provided in Appendix~\ref{append:annotation}.

We acknowledge that LLMs exhibit political biases~\cite{chalkidis-brandl-2024-llama}; however, we mitigate these by using RAG~\cite{lewis2020retrieval,ram2023context} and providing the URLs of the sources used for the timeline construction, so that users can verify the output themselves. 
Furthermore, we evaluated the system with fact-checkers from Full Fact, which is a signatory to the International Fact-Checkers Network code of principles (\url{https://ifcncodeofprinciples.poynter.org/the-commitments}) that stipulates that they need to be impartial in their work.

The release of our demo has been approved by the Ethics Review Committee\footnote{\url{https://www.cst.cam.ac.uk/local/policy/ethics}} at the Department of Computer Science and Technology, University of Cambridge, under the CC-BY-NC license.

\section*{Acknowledgements}
{We appreciate the program chairs, area chair, and reviewers from EMNLP 2025 System Demonstration for their insightful feedback.}
We would like to thank Chenxi Whitehouse and Tom Stafford for their contributions to an earlier version of this project.
This work is supported by the ERC grant \textsc{AVeriTeC} (GA 865958).
Full Fact’s work is funded by the JournalismAI Innovation Challenge, supported by the Google News Initiative.
Michael is further supported by the Engineering and Physical Sciences Research Council (grant number EP/Y009800/1), through funding from Responsible AI UK (KP0016). 


\bibliography{custom}

\begin{thebibliography}{34}
\providecommand{\natexlab}[1]{#1}

\bibitem[{Barbaresi(2021)}]{barbaresi-2021-trafilatura}
Adrien Barbaresi. 2021.
\newblock \href {https://doi.org/10.18653/v1/2021.acl-demo.15} {Trafilatura: {A} web scraping library and command-line tool for text discovery and extraction}.
\newblock In \emph{Proceedings of the 59th Annual Meeting of the Association for Computational Linguistics and the 11th International Joint Conference on Natural Language Processing: System Demonstrations}, pages 122--131, Online. Association for Computational Linguistics.

\bibitem[{Chalkidis and Brandl(2024)}]{chalkidis-brandl-2024-llama}
Ilias Chalkidis and Stephanie Brandl. 2024.
\newblock \href {https://aclanthology.org/2024.naacl-short.40} {Llama meets {EU}: Investigating the {E}uropean political spectrum through the lens of {LLM}s}.
\newblock In \emph{Proceedings of the 2024 Conference of the North American Chapter of the Association for Computational Linguistics: Human Language Technologies (Volume 2: Short Papers)}, pages 481--498, Mexico City, Mexico. Association for Computational Linguistics.

\bibitem[{Chen et~al.(2024{\natexlab{a}})Chen, Qin, Jiang, and Choi}]{chen2024large}
Ruirui Chen, Chengwei Qin, Weifeng Jiang, and Dongkyu Choi. 2024{\natexlab{a}}.
\newblock \href {https://doi.org/10.1609/AAAI.V38I16.29730} {Is a large language model a good annotator for event extraction?}
\newblock In \emph{Thirty-Eighth {AAAI} Conference on Artificial Intelligence, {AAAI} 2024, Thirty-Sixth Conference on Innovative Applications of Artificial Intelligence, {IAAI} 2024, Fourteenth Symposium on Educational Advances in Artificial Intelligence, {EAAI} 2014, February 20-27, 2024, Vancouver, Canada}, pages 17772--17780. {AAAI} Press.

\bibitem[{Chen et~al.(2024{\natexlab{b}})Chen, Liu, Yan, Bai, Zhong, Yang, Yang, Zhu, and Zhang}]{chen2024see}
Yulong Chen, Yang Liu, Jianhao Yan, Xuefeng Bai, Ming Zhong, Yinghao Yang, Ziyi Yang, Chenguang Zhu, and Yue Zhang. 2024{\natexlab{b}}.
\newblock \href {https://openreview.net/forum?id=18iNTRPx8c} {See what {LLM}s cannot answer: A self-challenge framework for uncovering {LLM} weaknesses}.
\newblock In \emph{First Conference on Language Modeling}.

\bibitem[{Costello and Thomson(2008)}]{costello2008election}
Rory Costello and Robert Thomson. 2008.
\newblock Election pledges and their enactment in coalition governments: A comparative analysis of ireland.
\newblock \emph{Journal of Elections, Public Opinion and Parties}, 18(3):239--256.

\bibitem[{Deng et~al.(2024)Deng, Schlichtkrull, and Vlachos}]{deng-etal-2024-document}
Zhenyun Deng, Michael Schlichtkrull, and Andreas Vlachos. 2024.
\newblock \href {https://doi.org/10.18653/v1/2024.acl-long.645} {Document-level claim extraction and decontextualisation for fact-checking}.
\newblock In \emph{Proceedings of the 62nd Annual Meeting of the Association for Computational Linguistics (Volume 1: Long Papers)}, pages 11943--11954, Bangkok, Thailand. Association for Computational Linguistics.

\bibitem[{Dupont et~al.(2019)Dupont, Bytzek, Steffens, and Schneider}]{dupont2019kind}
Julia~C Dupont, Evelyn Bytzek, Melanie~C Steffens, and Frank~M Schneider. 2019.
\newblock Which kind of political campaign messages do people perceive as election pledges?
\newblock \emph{Electoral Studies}, 57:121--130.

\bibitem[{Duval and P{\'e}try(2020)}]{duval2020citizens}
Dominic Duval and Fran{\c{c}}ois P{\'e}try. 2020.
\newblock Citizens’ evaluations of campaign pledge fulfillment in canada.
\newblock \emph{Party Politics}, 26(4):437--447.

\bibitem[{Fornaciari et~al.(2021)Fornaciari, Hovy, Naurin, Runeson, Thomson, and Adhikari}]{fornaciari-etal-2021-will}
Tommaso Fornaciari, Dirk Hovy, Elin Naurin, Julia Runeson, Robert Thomson, and Pankaj Adhikari. 2021.
\newblock \href {https://doi.org/10.18653/v1/2021.findings-acl.301} {{``}we will reduce taxes{''} - identifying election pledges with language models}.
\newblock In \emph{Findings of the Association for Computational Linguistics: ACL-IJCNLP 2021}, pages 3406--3419, Online. Association for Computational Linguistics.

\bibitem[{Gao et~al.(2023)Gao, Zhao, Yu, and Xu}]{gao2023exploring}
Jun Gao, Huan Zhao, Changlong Yu, and Ruifeng Xu. 2023.
\newblock \href {https://arxiv.org/abs/2303.03836} {Exploring the feasibility of chatgpt for event extraction}.
\newblock \emph{ArXiv preprint}, abs/2303.03836.

\bibitem[{Grattafiori et~al.(2024)Grattafiori, Dubey, Jauhri, Pandey, Kadian, Al-Dahle, Letman, Mathur, Schelten, Vaughan et~al.}]{grattafiori2024Llama}
Aaron Grattafiori, Abhimanyu Dubey, Abhinav Jauhri, Abhinav Pandey, Abhishek Kadian, Ahmad Al-Dahle, Aiesha Letman, Akhil Mathur, Alan Schelten, Alex Vaughan, and 1 others. 2024.
\newblock \href {https://arxiv.org/abs/2407.21783} {The llama 3 herd of models}.
\newblock \emph{ArXiv preprint}, abs/2407.21783.

\bibitem[{Guo et~al.(2022)Guo, Schlichtkrull, and Vlachos}]{guo2022survey}
Zhijiang Guo, Michael Schlichtkrull, and Andreas Vlachos. 2022.
\newblock \href {https://doi.org/10.1162/tacl_a_00454} {A survey on automated fact-checking}.
\newblock \emph{Transactions of the Association for Computational Linguistics}, 10:178--206.

\bibitem[{Hu et~al.(2022)Hu, Shen, Wallis, Allen{-}Zhu, Li, Wang, Wang, and Chen}]{hu2022lora}
Edward~J. Hu, Yelong Shen, Phillip Wallis, Zeyuan Allen{-}Zhu, Yuanzhi Li, Shean Wang, Lu~Wang, and Weizhu Chen. 2022.
\newblock \href {https://openreview.net/forum?id=nZeVKeeFYf9} {Lora: Low-rank adaptation of large language models}.
\newblock In \emph{The Tenth International Conference on Learning Representations, {ICLR} 2022, Virtual Event, April 25-29, 2022}. OpenReview.net.

\bibitem[{Hu et~al.(2024)Hu, Moon, and Ng}]{hu2024moments}
Qisheng Hu, Geonsik Moon, and Hwee~Tou Ng. 2024.
\newblock From moments to milestones: Incremental timeline summarization leveraging large language models.
\newblock In \emph{Proceedings of the 62nd Annual Meeting of the Association for Computational Linguistics (Volume 1: Long Papers)}, pages 7232--7246.

\bibitem[{Hurst et~al.(2024)Hurst, Lerer, Goucher, Perelman, Ramesh, Clark, Ostrow, Welihinda, Hayes, Radford et~al.}]{hurst2024gpt}
Aaron Hurst, Adam Lerer, Adam~P Goucher, Adam Perelman, Aditya Ramesh, Aidan Clark, AJ~Ostrow, Akila Welihinda, Alan Hayes, Alec Radford, and 1 others. 2024.
\newblock \href {https://arxiv.org/abs/2410.21276} {Gpt-4o system card}.
\newblock \emph{ArXiv preprint}, abs/2410.21276.

\bibitem[{Iqbal et~al.(2024)Iqbal, Wang, Wang, Georgiev, Geng, Gurevych, and Nakov}]{iqbal-etal-2024-openfactcheck}
Hasan Iqbal, Yuxia Wang, Minghan Wang, Georgi~Nenkov Georgiev, Jiahui Geng, Iryna Gurevych, and Preslav Nakov. 2024.
\newblock \href {https://doi.org/10.18653/v1/2024.emnlp-demo.23} {{O}pen{F}act{C}heck: A unified framework for factuality evaluation of {LLM}s}.
\newblock In \emph{Proceedings of the 2024 Conference on Empirical Methods in Natural Language Processing: System Demonstrations}, pages 219--229, Miami, Florida, USA. Association for Computational Linguistics.

\bibitem[{Konstantinovskiy et~al.(2021)Konstantinovskiy, Price, Babakar, and Zubiaga}]{konstantinovskiy2021claim}
Lev Konstantinovskiy, Oliver Price, Mevan Babakar, and Arkaitz Zubiaga. 2021.
\newblock \href {https://doi.org/10.1145/3412869} {Toward automated factchecking: Developing an annotation schema and benchmark for consistent automated claim detection}.
\newblock \emph{Digital Threats}, 2(2).

\bibitem[{Lewis et~al.(2020)Lewis, Perez, Piktus, Petroni, Karpukhin, Goyal, K{\"{u}}ttler, Lewis, Yih, Rockt{\"{a}}schel, Riedel, and Kiela}]{lewis2020retrieval}
Patrick S.~H. Lewis, Ethan Perez, Aleksandra Piktus, Fabio Petroni, Vladimir Karpukhin, Naman Goyal, Heinrich K{\"{u}}ttler, Mike Lewis, Wen{-}tau Yih, Tim Rockt{\"{a}}schel, Sebastian Riedel, and Douwe Kiela. 2020.
\newblock \href {https://proceedings.neurips.cc/paper/2020/hash/6b493230205f780e1bc26945df7481e5-Abstract.html} {Retrieval-augmented generation for knowledge-intensive {NLP} tasks}.
\newblock In \emph{Advances in Neural Information Processing Systems 33: Annual Conference on Neural Information Processing Systems 2020, NeurIPS 2020, December 6-12, 2020, virtual}.

\bibitem[{Liao et~al.(2023)Liao, Peng, Huang, Zhang, Li, Shu, and Xie}]{liao2023muser}
Hao Liao, Jiahao Peng, Zhanyi Huang, Wei Zhang, Guanghua Li, Kai Shu, and Xing Xie. 2023.
\newblock Muser: A multi-step evidence retrieval enhancement framework for fake news detection.
\newblock In \emph{Proceedings of the 29th ACM SIGKDD Conference on Knowledge Discovery and Data Mining}, pages 4461--4472.

\bibitem[{Liu et~al.(2019)Liu, Ott, Goyal, Du, Joshi, Chen, Levy, Lewis, Zettlemoyer, and Stoyanov}]{liu2019roberta}
Yinhan Liu, Myle Ott, Naman Goyal, Jingfei Du, Mandar Joshi, Danqi Chen, Omer Levy, Mike Lewis, Luke Zettlemoyer, and Veselin Stoyanov. 2019.
\newblock \href {https://arxiv.org/abs/1907.11692} {Roberta: A robustly optimized bert pretraining approach}.
\newblock \emph{ArXiv preprint}, abs/1907.11692.

\bibitem[{Liu et~al.(2024)Liu, Zhou, Zhu, Lian, Li, Dou, Lian, and Nie}]{liu2024information}
Zheng Liu, Yujia Zhou, Yutao Zhu, Jianxun Lian, Chaozhuo Li, Zhicheng Dou, Defu Lian, and Jian-Yun Nie. 2024.
\newblock Information retrieval meets large language models.
\newblock In \emph{Companion Proceedings of the ACM Web Conference 2024}, pages 1586--1589.

\bibitem[{Minard et~al.(2015)Minard, Speranza, Agirre, Aldabe, van Erp, Magnini, Rigau, and Urizar}]{minard2015semeval}
Anne-Lyse Minard, Manuela Speranza, Eneko Agirre, Itziar Aldabe, Marieke van Erp, Bernardo Magnini, German Rigau, and Rub{\'e}n Urizar. 2015.
\newblock \href {https://doi.org/10.18653/v1/S15-2132} {{S}em{E}val-2015 task 4: {T}ime{L}ine: Cross-document event ordering}.
\newblock In \emph{Proceedings of the 9th International Workshop on Semantic Evaluation ({S}em{E}val 2015)}, pages 778--786, Denver, Colorado. Association for Computational Linguistics.

\bibitem[{Panchendrarajan and Zubiaga(2024)}]{panchendrarajan2024claim}
Rrubaa Panchendrarajan and Arkaitz Zubiaga. 2024.
\newblock Claim detection for automated fact-checking: A survey on monolingual, multilingual and cross-lingual research.
\newblock \emph{Natural Language Processing Journal}, 7:100066.

\bibitem[{Qorib et~al.(2025)Qorib, Hu, and Ng}]{qorib2025just}
Muhammad~Reza Qorib, Qisheng Hu, and Hwee~Tou Ng. 2025.
\newblock Just what you desire: Constrained timeline summarization with self-reflection for enhanced relevance.
\newblock In \emph{Proceedings of the AAAI Conference on Artificial Intelligence}, volume~39, pages 25065--25073.

\bibitem[{Ram et~al.(2023)Ram, Levine, Dalmedigos, Muhlgay, Shashua, Leyton-Brown, and Shoham}]{ram2023context}
Ori Ram, Yoav Levine, Itay Dalmedigos, Dor Muhlgay, Amnon Shashua, Kevin Leyton-Brown, and Yoav Shoham. 2023.
\newblock \href {https://doi.org/10.1162/tacl_a_00605} {In-context retrieval-augmented language models}.
\newblock \emph{Transactions of the Association for Computational Linguistics}, 11:1316--1331.

\bibitem[{Sahnan et~al.(2025)Sahnan, Corney, Larraz, Zagni, Miguez, Xie, Gurevych, Churchill, Chakraborty, and Nakov}]{sahnan2025llmsautomatefactcheckingarticle}
Dhruv Sahnan, David Corney, Irene Larraz, Giovanni Zagni, Ruben Miguez, Zhuohan Xie, Iryna Gurevych, Elizabeth Churchill, Tanmoy Chakraborty, and Preslav Nakov. 2025.
\newblock \href {https://arxiv.org/abs/2503.17684} {Can llms automate fact-checking article writing?}

\bibitem[{Schlichtkrull et~al.(2024)Schlichtkrull, Chen, Whitehouse, Deng, Akhtar, Aly, Guo, Christodoulopoulos, Cocarascu, Mittal et~al.}]{schlichtkrull2024automated}
Michael Schlichtkrull, Yulong Chen, Chenxi Whitehouse, Zhenyun Deng, Mubashara Akhtar, Rami Aly, Zhijiang Guo, Christos Christodoulopoulos, Oana Cocarascu, Arpit Mittal, and 1 others. 2024.
\newblock The automated verification of textual claims (averitec) shared task.
\newblock In \emph{Proceedings of the Seventh Fact Extraction and VERification Workshop (FEVER)}, pages 1--26.

\bibitem[{Schlichtkrull et~al.(2023)Schlichtkrull, Guo, and Vlachos}]{schlichtkrull2023averitec}
Michael Schlichtkrull, Zhijiang Guo, and Andreas Vlachos. 2023.
\newblock \href {http://papers.nips.cc/paper\_files/paper/2023/hash/cd86a30526cd1aff61d6f89f107634e4-Abstract-Datasets\_and\_Benchmarks.html} {Averitec: {A} dataset for real-world claim verification with evidence from the web}.
\newblock In \emph{Advances in Neural Information Processing Systems 36: Annual Conference on Neural Information Processing Systems 2023, NeurIPS 2023, New Orleans, LA, USA, December 10 - 16, 2023}.

\bibitem[{Seki et~al.(2024)Seki, Shu, Lhuissier, Lee, Kang, Day, and Chen}]{seki2024ml}
Yohei Seki, Hakusen Shu, Ana{\"\i}s Lhuissier, Hanwool Lee, Juyeon Kang, Min-Yuh Day, and Chung-Chi Chen. 2024.
\newblock \href {https://arxiv.org/abs/2411.04473} {Ml-promise: A multilingual dataset for corporate promise verification}.
\newblock \emph{ArXiv preprint}, abs/2411.04473.

\bibitem[{Turk et~al.(2025)Turk, Khan, and Kosseim}]{turk2025clac}
Nawar Turk, Eeham Khan, and Leila Kosseim. 2025.
\newblock Clac at semeval-2025 task 6: A multi-architecture approach for corporate environmental promise verification.
\newblock \emph{arXiv preprint arXiv:2505.23538}.

\bibitem[{Yang et~al.(2024)Yang, Rao, Chen, Guo, Zhang, Yang, and Zhang}]{yang2024rag}
Diji Yang, Jinmeng Rao, Kezhen Chen, Xiaoyuan Guo, Yawen Zhang, Jie Yang, and Yi~Zhang. 2024.
\newblock \href {https://doi.org/10.1145/3626772.3657760} {{IM-RAG:} multi-round retrieval-augmented generation through learning inner monologues}.
\newblock In \emph{Proceedings of the 47th International {ACM} {SIGIR} Conference on Research and Development in Information Retrieval, {SIGIR} 2024, Washington DC, USA, July 14-18, 2024}, pages 730--740. {ACM}.

\bibitem[{Yoon et~al.(2024)Yoon, Jung, Yoon, and Park}]{yoon-etal-2024-hero}
Yejun Yoon, Jaeyoon Jung, Seunghyun Yoon, and Kunwoo Park. 2024.
\newblock \href {https://doi.org/10.18653/v1/2024.fever-1.15} {{H}er{O} at {AV}eri{T}e{C}: The herd of open large language models for verifying real-world claims}.
\newblock In \emph{Proceedings of the Seventh Fact Extraction and VERification Workshop (FEVER)}, pages 130--136, Miami, Florida, USA. Association for Computational Linguistics.

\bibitem[{Zhang et~al.(2024)Zhang, Guo, and Vlachos}]{zhang2024we}
Caiqi Zhang, Zhijiang Guo, and Andreas Vlachos. 2024.
\newblock Do we need language-specific fact-checking models? the case of chinese.
\newblock In \emph{Proceedings of the 2024 Conference on Empirical Methods in Natural Language Processing}, pages 1899--1914.

\bibitem[{Zhang et~al.(2023)Zhang, Li, Cui, Cai, Liu, Fu, Huang, Zhao, Zhang, Chen et~al.}]{zhang2023siren}
Yue Zhang, Yafu Li, Leyang Cui, Deng Cai, Lemao Liu, Tingchen Fu, Xinting Huang, Enbo Zhao, Yu~Zhang, Yulong Chen, and 1 others. 2023.
\newblock \href {https://arxiv.org/abs/2309.01219} {Siren's song in the ai ocean: a survey on hallucination in large language models}.
\newblock \emph{ArXiv preprint}, abs/2309.01219.

\end{thebibliography}

\newpage
\appendix

\section{Implementation Details}
\label{append:prompt}
\subsection{Evidence Retrieval}\label{append:prompt_question_generation}

Following~\citet{yoon-etal-2024-hero}, we index the training data from AVeriTeC~\cite{schlichtkrull2023averitec} and retrieve the top-10 most similar question-evidence pairs to the input pledge from the training corpus using BM25.
These top-10 question-evidence pairs are then used as the ICL examples.
In particular, the prompt is as follow:
\begin{quote}
Your task is to generate a question based on the given claim and evidence. The question should clarify the relationship between the evidence and the claim.

\{\texttt{ICL\_examples}\}

Now, generate a question that links the following claim and evidence:

{Claim:} \{\texttt{pledge\_claim}\}\\
{Evidence:} \{\texttt{sentence\_evidence\}}
\end{quote}
We use \texttt{Meta-Llama-3.1-8B-Instruct} with a temperature of 0.6 and top-$p$ of 0.9. We generate one question per evidence sentence.

For Google Custom Search, we set the geographic scope to the UK due to our focus on the UK election pledges.
In practice, we set the iterative evidence retrieval to two rounds, to balance a good result in practice and our budgets.

\subsection{Timeline Construction}\label{append:prompt_eevent_generation}

We use the below prompt for event description generation and timestamp identification:
\begin{quote}
Please only summarize events that are useful for verifying the pledge, and their dates in the JSON format.

\{\texttt{ICL\_examples}\}

Please only summarize events that are useful for verifying the pledge: \texttt{\{pledge\}}, and their dates in the JSON format.

{Input:}

Title: \{\texttt{document\_title}\}\\
Date: \{\texttt{document\_date}\}\\
Article: \{\texttt{document\_text}\}

{Output:}
\end{quote}
Please note that we use \texttt{GPT-4o} for experiments, and constrain the output (including the outputs of the ICL pairs) in the JSON format, for example:

\begin{quote}\texttt{\{\\
  "events": [\\
    \{\\
      "event": "Home Secretary Yvette Cooper announces new measures to boost Britain's border security, including the recruitment of up to 100 new specialist intelligence and investigation officers at the National Crime Agency (NCA).",\\
      "date": "2024-08-21"\\
    \},\\
    \{\\
      "event":\\ "Announcement of a major surge in immigration enforcement and returns activity to achieve the highest rate of removals of those with no right to be in the UK since 2018.",\\
      "date": "2024-08-21"\\
    \},\\
    \{\\
    ...\\
    \}]\\
\}
}
\end{quote}

We use 2 ICL examples to balance the length constraint and model efficiency.
We set the top-$p$ and temperature to 0.

\subsection{Relevant Event Identification}\label{append:prompt_relevant_event}
We use the below prompt for identifying relevant events:
\begin{quote}
You are given a pledge, the pledge speaker, and the date of when the pledge is made, and a key event summarized from an online article along with the date of when the event happens. Your task is to determine whether this event summary is useful to track the fulfilment of this pledge. 

Yes: The summary presents developments or actions that demonstrate progress (or lack thereof) towards fulfilling the pledge. It helps evaluate whether the pledge is on track or not.

No: The summary only provides background or contextual information, but no progress information for evaluating the fulfilment of the pledge; Or the summary is less than or not related to the pledge.

Below are examples:

\{\texttt{ICL\_examples}\}

Now, please assign a label to the below instance.

Input:

Pledge: \{\texttt{pledge}\}\\
Event summary: \{\texttt{event}\}. 
(Event Date: \{\texttt{event\_date}\})

Output:
\end{quote}
The model is expected to return \texttt{Yes} or \texttt{No}, and we also log the log-probability of the first predicted token to support confidence-based ranking.

We use at most 50 ICL examples. In particular, in our demo system, if we are checking a suggested pledge, we use their corresponding annotated data; otherwise, we randomly select instances from all annotated data.
We set the top-$p$ and temperature to 0.






\section{Fact-checkers' Background}\label{append:annotation}

Both of the fact-checkers involved in this study (Nasim and Josh) are native English speakers, educated to postgraduate level. 
One has worked as a fact-checker for two years and overall as a trained journalist for seven years, while the other has worked as a journalist for eight years and in fact-checking for several months.

\section{Setup of Google Search and GPT-4o for Real-world Evaluation}\label{append: gpt_4o_for_evidence_retrieval}

We use \texttt{GPT-4o} with the tool of \texttt{web\_search}.
We set the location as the UK (\texttt{GB} in \texttt{GPT-4o}), and the \texttt{search\_context\_size} as \texttt{high}.
We use the same request for initial searching as the input, and use the below prompt to inform the model of the time range:
\begin{quote}
    Please find the recent online articles (from \texttt{\{time\_start\}} to \texttt{\{time\_end\}}) that can help monitor the fulfilment of the pledge. List only the article URLs, ordered by their usefulness and relevance (most useful and relevant first), one per line.

    \texttt{\{pledge\}}
\end{quote}

Similarly, we use the same API for \textsc{PledgeTracker} to call Google Search using the same parameters, and collect the top-10 retrieved results based on their prominence.

To ensure the retrieved URLs are within the correct time range, we further filter all URLs by examining their metadata, and use the useful URLs for evaluation.

\end{document}